# Training Deep Convolutional Neural Networks with Resistive Cross-Point Devices

**Authors:** Tayfun Gokmen,* O. Murat Onen, Wilfried Haensch

**Affiliations**
IBM T.J. Watson Research Center, Yorktown Heights, NY 10598 USA
*Correspondence to: tgokmen@us.ibm.com

**Abstract**
In a previous work we have detailed the requirements to obtain a maximal performance benefit by implementing fully connected deep neural networks (DNN) in form of arrays of resistive devices for deep learning. This concept of Resistive Processing Unit (RPU) devices we extend here towards convolutional neural networks (CNNs). We show how to map the convolutional layers to RPU arrays such that the parallelism of the hardware can be fully utilized in all three cycles of the backpropagation algorithm. We find that the noise and bound limitations imposed due to analog nature of the computations performed on the arrays effect the training accuracy of the CNNs. Noise and bound management techniques are presented that mitigate these problems without introducing any additional complexity in the analog circuits and can be addressed by the digital circuits. In addition, we discuss digitally programmable update management and device variability reduction techniques that can be used selectively for some of the layers in a CNN. We show that combination of all those techniques enables a successful application of the RPU concept for training CNNs. The techniques discussed here are more general and can be applied beyond CNN architectures and therefore enables applicability of RPU approach for large class of neural network architectures.

# INTRODUCTION

Deep neural network (DNN) [1] based models demonstrated unprecedented accuracy, some of which exceeding human level performance, in cognitive tasks such as object recognition [2] [3] [4] [5], speech recognition [6], and natural language processing [7]. These accomplishments are made possible thanks to the advances in computing architectures and the availability of large amounts of labelled training data. Furthermore, network architectures are adjusted to take advantage of data properties such as spatial or temporal correlation. For instance, convolutional neural networks (CNN) provide superior results for image recognition and recurrent neural networks (RNN) in speech and natural language processing. Therefore, the application space of the traditional fully connected deep learning network is diminishing. In a recent paper we have introduced the concept of a restive processing unit (RPU) as an architecture solution for fully connected DNN. We will now show that the RPU concept is equally applicable for CNN.

Training large DNNs is an extremely computationally intensive task that may take weeks even on distributed parallel computing frameworks utilizing many computing nodes [8] [9] [10]. There have been many attempts to accelerate DNN training by designing and using specialized hardware such as GPUs [11] [12], FPGAs [13], or ASICs [14] that rely on conventional CMOS-technology. All of these approaches share the common objective of packing larger number of computing units into a fixed area and power budget by using optimized multiply and add units so that acceleration over a CPU can be achieved. Although various microarchitectures and data formats are considered for different accelerator designs [15] [16] [17], all of these digital approaches use a similar underlying transistor technology and therefore the acceleration factors will eventually be limited due to scaling limitations.

In order to achieve even larger acceleration factors beyond conventional CMOS, novel nano-electronic device concepts based on non-volatile memory (NVM) technologies [18], such as phase change memory (PCM) [19], resistive random access memory (RRAM) [20], and memristors [21] [22] [23] have been explored for implementing DNN training. Acceleration factors ranging from $25X - 2000X$ [24] [25] [26] compared to the conventional CPU/GPU based approaches and significant reduction in power and area have been proposed. However, for these bottom-up approaches the acceleration factors are still limited by device specifications intrinsic to their application as non-volatile memory (NVM) cells. Instead, using a top-down approach it is possible to develop a new class of devices, so called Resistive Processing Unit (RPU) devices [27] that are free from these limitations, and therefore can promise ultimate accelerations factors of $30,000X$ while providing power efficiency of $84,000\ GigaOps/s/W$.

The concept of using resistive cross-point device arrays [27] [28] [29] as DNN accelerators have been tested up to some extend by performing simulations using fully connected neural networks. The effect of various device features and system parameters on training performance is evaluated to derive the device and system level specifications for a successful implementation of an accelerator chip for DNN compute efficient training [27] [30]. It is shown that resistive devices, that are analog in nature, need to respond symmetrically in up and down conductance changes provided the same but opposite pulse stimulus. Indeed, these specifications differ significantly from parameters typically used for memory elements and therefore require a systematic search for new physical mechanisms, materials and device designs to realize an ideal resistive element for DNN training. It is important to note that, however, these resistive cross-point arrays perform the multiply and add in the analog domain in contrast to the CMOS based digital approaches. Therefore, it is not clear yet whether the proposed device specifications that are sufficient to train a fully connected neural network generalize to a more general set of network architectures, and hence requires further validation of their applicability to a broader range of networks.

**Fully Connected Neutral Networks**

Deep fully connected neural networks are composed of stacking of multiple fully connected layers such that the signal propagates from input layer to output layer by going through series of linear and non-linear transformations. The whole network expresses a single differentiable error function that maps the input data on to class scores at the output layer. Most commonly the network is trained with simple stochastic gradient decent (SGD), in which the error gradient with respect to each parameter is calculated using the backpropagation algorithm [31].

The backpropagation algorithm is composed of three cycles, forward, backward and weight update that are repeated many times until a convergence criterion is met. For a single fully connected layer where $N$ inputs neurons are connected to $M$ output (or hidden) neurons, the forward cycle involve computing a vector-matrix multiplication ($\boldsymbol{y} = \boldsymbol{Wx}$) where the vector $\boldsymbol{x}$ of length $N$ represents the activities of the input neurons and the matrix $\boldsymbol{W}$ of size $M \times N$ stores the weight values between each pair of input and output neurons. The resulting vector $\boldsymbol{y}$ of length $M$ is further processed by performing a non-linear activation on each of the elements and then passed to the next layer. Once the information reaches to the final output layer, the error signal is calculated and backpropagated through the network. The backward cycle on a single layer also involves a vector-matrix multiplication on the transpose of the weight matrix ($\boldsymbol{z} = \boldsymbol{W^T \delta}$), where the vector $\boldsymbol{\delta}$ of length $M$ represents the error calculated by the output neurons and the vector $\boldsymbol{z}$ of length $N$ is further processed using the derivative of neuron non-linearity and then passed down to the previous layers. Finally, in the update cycle the weight matrix $\boldsymbol{W}$ is updated by performing an outer product of the two vectors that are used in the forward and the backward cycles and usually expressed as $\boldsymbol{W} \leftarrow \boldsymbol{W} + \eta\,(\boldsymbol{\delta x^T})$ where $\eta$ is a global learning rate.

**Mapping Fully Connected Layers to Resistive Device Arrays**

All of the above operations performed on the weight matrix $\boldsymbol{W}$ can be implemented with a 2D crossbar array of two-terminal resistive devices with $M$ rows and $N$ columns where the stored conductance values in the crossbar array form the matrix $\boldsymbol{W}$. In the forward cycle, input vector $\boldsymbol{x}$ is transmitted as voltage pulses through each of the columns and resulting vector $\boldsymbol{y}$ can be read as current signals from the rows [32]. Similarly, when voltage pulses are supplied from the rows as an input in the backward cycle, then a vector-matrix product is computed on the transpose of the weight matrix $\boldsymbol{W^T}$. Finally, in the update cycle voltage pulses representing vectors $\boldsymbol{x}$ and $\boldsymbol{\delta}$ are simultaneously supplied from the columns and the rows. At this setting each cross-point device performs a local multiplication and summation operation by processing the voltage pulses coming from the column and the row and hence achieving an incremental weight update.

All three operating modes described above allow the arrays of cross-point devices that constitute the network to be active in all three cycles and hence enable a very efficient implementation of the backpropagation algorithm. Because of their local weight storage and processing capability these resistive cross-point devices are called Resistive Processing Unit (RPU) devices [27]. An array of RPU devices can perform the operations involving the weight matrix $\boldsymbol{W}$ locally and in parallel, and hence achieves $O(1)$ time complexity in all three cycles independent of the array size.

Here, we extend the RPU device concept towards convolutional neural networks (CNNs). First we show how to map the convolutional layers to RPU device arrays such that the parallelism of the hardware can be fully utilized in all three cycles of the backpropagation algorithm. Next, we show that identical RPU device specifications hold for CNNs. Our study shows, however, that CNNs are more sensitive to noise and bounds due to analog nature of the computations on RPU arrays. We discuss noise and bound management techniques that mitigate these problems without introducing any additional complexity in the analog circuits and can be addressed by the digital circuits. In addition, we discuss digitally programmable update management and device variability reduction techniques that can be used selectively for some of the layers in a CNN. We show that combination of all those techniques enables a successful application of the RPU concept for training CNNs; and a network trained with all the RPU device imperfections can yield a classification error indistinguishable from a network trained with high precision floating point numbers.

## MATERIALS AND METHODS

### Convolutional Layers

The input to a convolutional layer can be an image or an output of the previous convolutional layer and is generally considered as a volume with dimensions of $(n, n, d)$ with a width and height of $n$ pixels and a depth of $d$ channels corresponding to different input components (e.g. red, green and blue components of an image) as illustrated in Figure 1A. The kernels of a convolutional layer are also a volume that is spatially small along the width and height, but extends through the full depth of the input volume with dimensions of $(k, k, d)$. During the forward cycle, each kernel slides over the input volume across the width and height and a dot product is computed between the parameters of the kernels and the input pixels at any position. Assuming no zero padding and single pixel sliding (stride is equal to one), this 2D convolution operation results in a single output plane with dimensions $((n - k + 1), (n - k + 1), 1)$ per kernel. Since there exists $M$ different kernels, output becomes a volume with dimensions $((n - k + 1), (n - k + 1), M)$ and is passed to following layers for further processing. During the backward cycle of a convolutional layer similar operations are performed but this time the spatially flipped kernels slide over the error signals that are backpropagated from the upper layers. The error signals form a volume with the same dimensions of the output $((n - k + 1), (n - k + 1), M)$. The results of this backward convolution are organized to a volume with dimensions $(n, n, d)$ and are further backpropagated for error calculations in the previous layers. Finally, in the update cycle, gradient with respect to each parameter is computed by convolving the input volume with the error volume used in the forward and backward cycles, respectively. This gradient information, which has the same dimensions as the kernels, is added to the kernel parameters after scaled with a learning rate.

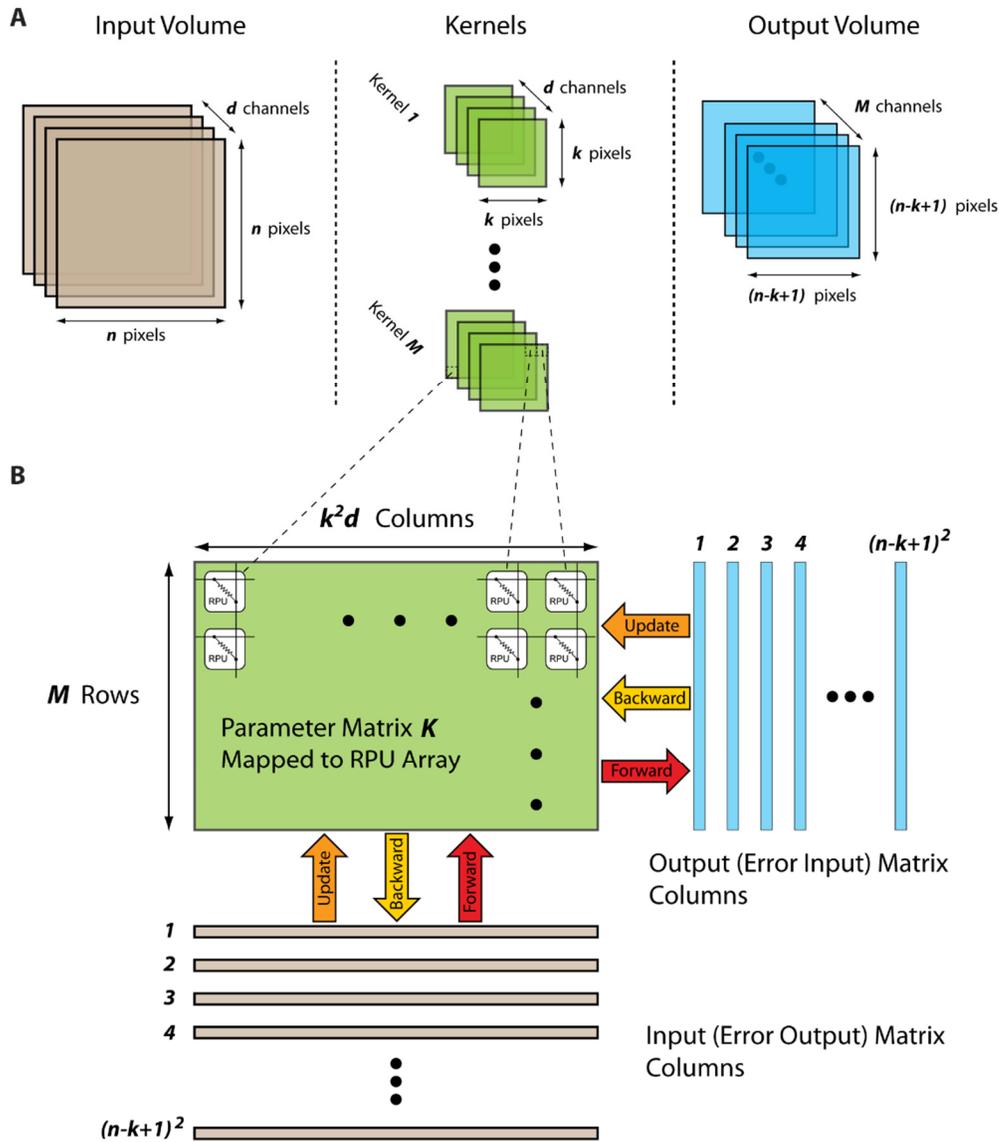

**Figure 1. (A)** Schematics of a convolutional layer showing the input volume, kernels, and the output volume. **(B)** Schematics of a mapped convolutional layer to an RPU array showing the input and output matrixes and their propagation through the kernel matrix during the forward, backward and the update cycles.

## Mapping Convolutional Layers to Resistive Device Arrays

For an efficient implementation of a convolutional layer using an RPU array, all the input/output volumes as well as the kernel parameters need to be rearranged in a specific way. The convolution operation essentially performs a dot product between the kernel parameters and a local region of the input volume and hence can be formulated as a matrix-matrix multiply. By collapsing the parameters of a single kernel to a column vector of length $k^2d$ and stacking all $M$ different kernels as separate rows, a parameter matrix $K$ of size $M \times k^2d$ is formed that stores all of the trainable parameters associated a single convolutional layer as shown in Figure 1B. After this rearrangement, in the forward cycle the outputs corresponding to a specific location along the width and height can be calculated by performing a vector-matrix multiplication $y = Kx$, where the vector $x$ of length $k^2d$ is a local region in the input volume and vector $y$ of length $M$ has all the results along the depth of the output volume. By repeating this vector-matrix multiplication for different local regions, the full volume of the output map can be computed. Indeed, this repeated vector-matrix multiplication is equivalent to a matrix-matrix multiplication $Y = KX$, where the matrix $X$ with dimensions $k^2d \times (n - k + 1)^2$ has the input neuron activities with some repetition and resulting matrix $Y$ with dimensions $M \times (n - k + 1)^2$ has all the results corresponding to the output volume. Similarly, using the transpose of the parameter matrix, the backward cycle of a convolutional layer can also be expresses as a matrix-matrix multiplication $Z = K^T D$, where the matrix $D$ with dimensions $M \times (n - k + 1)^2$ has the error signals corresponding to an error volume. Furthermore, in this setting the update cycle also simplifies to a matrix multiplication where the gradient information for the whole parameter matrix $K$ can be computed using matrices $X$ and $D$, and the update rule can be written as $K \leftarrow K + \eta(DX^T)$.

The rearrangement of the trainable parameters to a single matrix $K$ by flattening of the kernels enables an efficient implementation of a convolutional layer using an RPU array. After this rearrangement, all the matrix operations performed on $K$ can be computed as a series of vector operations on an RPU array. Analogous to the fully connected layers, matrix $K$ is mapped to an RPU array with $M$ rows and $k^2d$ columns as shown in Figure 1B. In the forward cycle, the input vector corresponding to a single column in $X$ is transmitted as voltage pulses from the columns and the results are read from the rows. Repetition of this operation for all $(n - k + 1)^2$ columns in $X$ completes all the computations required for the forward cycle. Similarly, in the backward cycle the input vector corresponding to a single column in $D$ is serially fed to the rows of the array. The update rule shown above can be viewed as a series of updates that involves computing an outer product between two columns from $X$ and $D$. This can be achieved by serially feeding the columns of $X$ and $D$ simultaneously to the RPU array. During the update cycle each RPU device performs a series of local multiplication and summation operations and hence calculates the product of the two matrixes.

We note that for a single input the total number of multiplication and summation operations that need to be computed in all three cycles for a convolutional layer is $Mk^2d(n - k + 1)^2$ and this number is independent of the method of computation. The proposed RPU mapping described above achieves this number as follows: Due to the inherent parallelism in the RPU array $Mk^2d$ operations are performed simultaneously for each vector operation performed on the array. Since there are $(n - k + 1)^2$ many vector operations performed serially on the array total number of computations are justified. Alternatively, one can view that there exists $Mk^2d$ trainable parameters and each parameter is used $(n - k + 1)^2$ times thanks to the parameter sharing in a convolution layer. Since, each RPU device in an array can perform a single computation at any given time, parameter sharing is achieved by accessing the array $(n - k + 1)^2$ times. For fully connected layers each weight is used only once and therefore all the computations can be completed with single vector operation on the array.

The end result of mapping a convolutional layer onto the RPU array is very similar to the mapping of a fully connected layer and therefore does not change the fundamental operations performed on the array. We also emphasize that the described convolutional layer with no zero padding and single pixel sliding is only used for illustration purposes. However, the proposed mapping is more general and can be applied to convolutional layers with zero padding, strides larger than a single pixel, dilated convolutions or convolutions with non-square inputs or kernels.

**RESULTS**

In order to test the validity of this method we performed deep neural network training simulations for the MNIST dataset using a CNN architecture similar to LeNet-5 [33]. It comprises of two convolutional layers with $5 \times 5$ kernels and hyperbolic tangent ($tanh$) activation functions. The first layer has 16 kernels while the second layer has 32 kernels. Each convolutional layer is followed by a subsampling layer that implements the max pooling function over non-overlapping pooling windows of size $2 \times 2$. The output of the second pooling layer, consisting of 512 neuron activations, feeds into a fully connected layer consisting of 128 $tanh$ neurons, which is then connected into a 10-way $softmax$ output layer. Training is performed repeatedly using a mini-batch size of unity for all 60,000 images in the training dataset which constitutes a single training epoch. Learning rate of $\eta = 0.01$ is used throughout the training for all 30 epochs.

Following the proposed mapping above, the trainable parameters (including the biases) of this architecture are stored in 4 separate arrays with dimensions of $16 \times 26$ and $32 \times 401$ for the first two convolutional layers, and, $128 \times 513$ and $10 \times 129$ for the following two fully connected layers. We name these arrays as $K_1$, $K_2$, $W_3$ and $W_4$, where the subscript denotes the layer's location and $K$ and $W$ is used for convolutional and fully connected layers, respectively. When all four arrays are considered as simple matrices and the operations are performed with floating point (FP) numbers, the network achieves a classification error of 0.8% on the test data. This is the FP-baseline model that we compare against the RPU based simulations for the rest of the paper.

**RPU Baseline Model**

The influence of various RPU device features, variations, and non-idealities on the training accuracy of a deep fully connected network have been tested by Ref [27]. We follow the same methodology here and as a baseline for all our RPU models we use the device specifications that resulted in an acceptable test error on the fully connected network.

The RPU-baseline model uses the stochastic update scheme where the numbers that are encoded from neurons ($x_i$ and $\delta_j$) are translated to stochastic bit streams so that each RPU device can perform a stochastic multiplication [34] [35] [36] by performing simple coincidence detection as illustrated by Figure 2. In this update scheme the expected weight change can be written as

$$\mathbb{E}(\Delta w_{ij}) = BL\,\Delta w_{min}(C_x x_i)(C_\delta \delta_j) \qquad (1)$$

where $BL$ is the length of the stochastic bit stream, $\Delta w_{min}$ is the change in the weight value due to a single coincidence event, $C_x$ and $C_\delta$ are the gain factors used during the stochastic translation for the columns and the rows, respectively. The RPU-baseline has $BL = 10$, $C_x = C_\delta = \sqrt{\eta/(BL\,\Delta w_{min})} = 1.0$ and $\Delta w_{min} = 0.001$. The change in weight values is achieved by conductance change in RPU devices; therefore, in order to capture device imperfections, $\Delta w_{min}$ is assumed to have cycle-to-cycle and device-to-device variations of 30%. The fabricated RPU devices may also show different amounts of change to positive and negative weight updates. This is taken into account by using separate $\Delta w_{min}^+$ for the positive updates and $\Delta w_{min}^-$ for the negative updates for each RPU device. The average value of the ratio $\Delta w_{min}^+/\Delta w_{min}^-$ among all devices is assumed to be unity as this can be achieved by a global adjustment of the voltage pulse durations/heights. However, device-to-device mismatch is unavoidable and therefore 2% variation is introduced for this parameter. To take conductance saturation into account, which is expected to be present in realistic RPU devices, the bounds on the weights values, $|w_{ij}|$, is assumed be 0.6 on average with a 30% device-to-device variation. We did not introduce any non-linearity in the weight update as this effect is shown to be not important as long as the updates are reasonably balanced (symmetric) between up and down changes. During the forward and backward cycles the vector-matrix multiplications performed on an RPU array are prone to analog noise and signal saturation due to the peripheral circuitry. The array operations including the input and output signals are illustrated in Figure 2 for these two cycles. Because of the analog current integration during the measurement time ($t_{meas}$), the output voltage ($V_{out}$) generated over the integrating capacitor ($C_{int}$) will have noise contributions coming from various sources. These noise sources are taken into account by introducing an additional Gaussian noise, with zero mean and standard deviation of $\sigma = 0.06$, to the results of vector-matrix multiplications computed on an RPU array. In addition these results are bounded to a value of $|\alpha| = 12$ to account for a signal saturation on the output voltage corresponding to a supply voltage on the op-amp. Table 1 summarizes all of the RPU-baseline model parameters used in our simulations that are also consistent with the specifications derived by Ref [27].

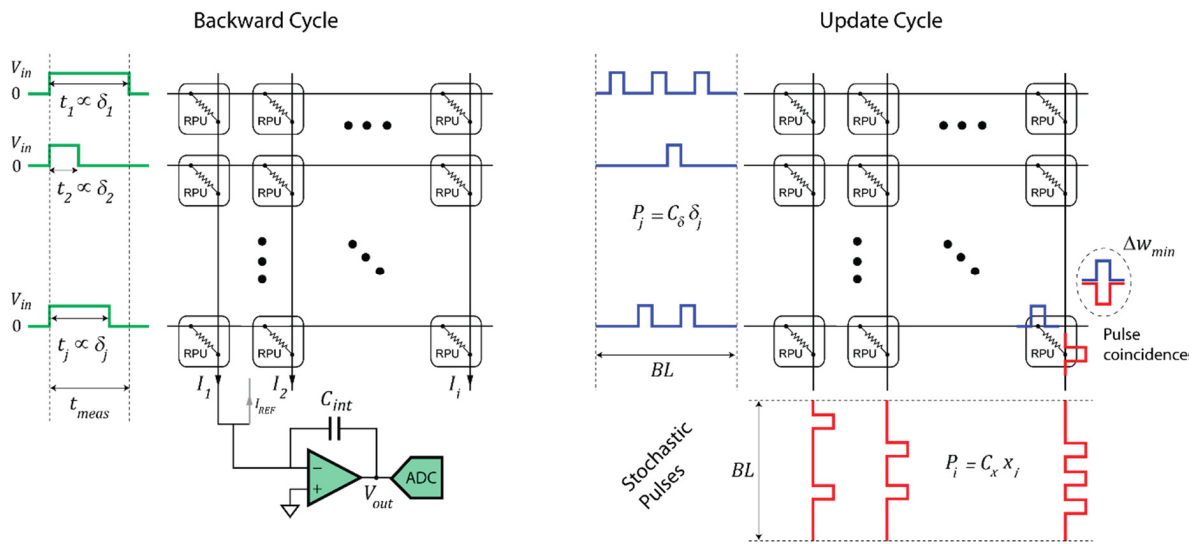

**Figure 2.** Schematics of an RPU array operation during the backward and update cycles. The forward cycle operations are identical to the backward cycle operations except the inputs are supplied from the columns and the outputs are read from the rows.

**Table 1. Summary of the RPU-baseline model parameters**

| BL | $C_x, C_\delta$ | $\Delta w_{min}$ | | | $\Delta w_{min}^+ / \Delta w_{min}^-$ | | $|w_{ij}|$ | | $\sigma$ | $|\alpha|$ |
|---|---|---|---|---|---|---|---|---|---|---|
| | | Average | Device to device variation | Cycle-to-cycle variation | Average | Device-to-device variation | Average | Device-to-device variation | Analog Noise | Signal Bound |
| 10 | 1.0 | 0.001 | 30% | 30% | 1.0 | 2% | 0.6 | 30% | 0.06 | 12 |

The CNN training results for various RPU variations are shown in Figure 3A. Interestingly, the RPU-baseline model of Table 1 performs poorly and only achieves a test error between 10% and 20% as shown by the black curve. Not only is this value significantly higher than the FP-baseline value of 0.8% but also higher than the 2.3% error rate achieved with the same RPU model for a fully connected network on the same dataset. Our investigation shows that this test error is due to simultaneous contributions of analog noise introduced during the backward cycle and signal bounds introduced in the forward cycle only on the final RPU array, $W_4$. As shown by the green curve, the model without analog noise in the backward cycle and infinite bounds on $W_4$ reaches a respectable test error of about 1.5%. When we only eliminate the noise while keeping the bounds, the model can follow a modest training up to about $8^{th}$ epoch but then the error rate suddenly increases and reaches a value about 10%. Similarly, if we only eliminate the bounds while keeping the noise, the model, shown by the red curve, performs poorly and the error rate stays around 10% level.

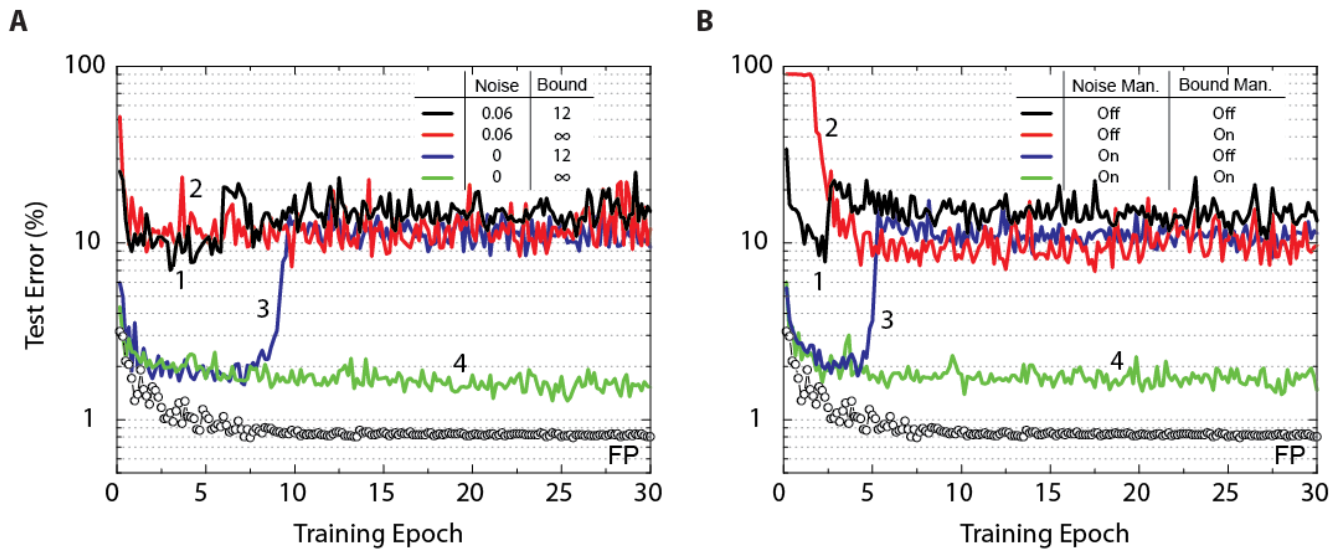

**Figure 3.** Test error of CNN with the MNIST dataset. Open white circles correspond to the model with the training performed using the floating point (FP) numbers. **(A)** Lines with different colors correspond to RPU-baseline models with different noise terms in the backward cycle and signals bounds on the last classification layer as given by the legend. **(B)** All lines marked with different colors correspond to RPU-baseline models including the noise and the bound terms; however, the noise management and the bound management techniques are applied selectively as given by the legend.

## Noise and Bound Management Techniques

It is clear that the noise in the backward cycle and the signal bounds on the output layer need to be addressed for successful application of RPU approach to CNN training. The complete elimination of analog noise and signal bounds is not realistic for real hardware implementation of RPU arrays. Designing very low noise read circuity with very large signal bounds is not an option because it will introduce unrealistic area and power constraints on the analog circuits. Below we describe noise and bound management techniques that can easily be implemented in the digital domain without changing the design considerations of RPU arrays and the supporting analog peripheral circuits.

During a vector-matrix multiplication on an RPU array, the input vector ($x$ or $\delta$) is transmitted as voltage pulses with a fixed amplitude and tunable durations as illustrated by Figure 2. In a naive implementation, the maximal pulse duration represents unity ($t_{meas} \rightarrow 1$), and all pulse durations are scaled accordingly depending on the values of $x_i$ or $\delta_j$. This scheme works optimally for the forward cycle with $tanh$ (or $sigmoid$) activations, as all $x_i$ in $x$ including a bias term are between $[-1,1]$. However, this assumption becomes problematic for the backward cycle, as there are not any guarantees for the range of the error signals in $\delta$. For instance, all $\delta_j$ in $\delta$ may become significantly smaller than unity ($\delta \ll 1$) as the training progresses and the classification error gets smaller. In this scenario the results of a vector-matrix multiplication in the backward cycle, as shown by Eq (2) below,

$$z = W^T \delta + \sigma \qquad (2)$$

are dominated by the noise term $\sigma$, as the signal term $W^T \delta$ does not generate enough voltage at the output. This is indeed why the noise introduced in the backward cycle brings the learning to a halt at around 10% error rate as shown by models in Figure 3A.

In order to get better signal at the output even when $\delta \ll 1$, we divide all $\delta_j$ in $\delta$ to the maximum value $\delta_{max}$ before the vector-matrix multiplication is performed on an RPU array. We note that this division operation is performed on digital circuits and ensures that at least one signal at the input of an RPU array exists for the whole integration time corresponding to unity. After the results of the vector-matrix multiplication is read from an RPU array and converted back to digital signals, we rescale the results by the same amount $\delta_{max}$. In this noise management scheme, the results of a vector-matrix multiplication can be written as

$$z = \left[ W^T \left[ \frac{\delta}{\delta_{max}} \right] + \sigma \right] \delta_{max}. \qquad (3)$$

The result, $z = W^T \delta + \sigma \delta_{max}$, effectively reduces the impact of noise significantly for small error rates $\delta_{max} \ll 1$. This noise management scheme allows to propagate error signals that are arbitrarily small and maintains a fixed signal to noise ratio independent of the range of numbers in $\delta$.

In addition to the noise, results of a vector-matrix multiplication will be bounded by $|\alpha|$ term that corresponds to a maximum allowed voltage during the integration time. This value $|\alpha| = 12$ is not critical while calculating activations for hidden layers with $tanh$ (or $sigmoid$) non-linearity because the error introduced due to the bound is negligible for a value that is otherwise much larger. However, for the output layer with $softmax$ (or $ReLU$) activations the error introduced due to the bound may become significant. For instance, if there exist two outputs that are above the bounded value, they would be treated equally strong and the classification task would result an equally probable two classes even if one of the outputs is significantly larger than the other. This results in a significant error (major information loss) in estimating the class label and hence limits the performance of the network. Similar to the noise, the bounded signals start to become an issue for later stages of the training as the network starts to perform good test results and the decision boundary between classes become more distinct. As shown by the blue curve in Figure 3A, at the beginning of the training the network successfully learns and test errors as low as 2% can be achieved; however, right around $8^{th}$ epoch the misleading error signals due to signal bounding forces the network to learn unwanted features and hence the error rate suddenly increases.

In order to eliminate the error introduced due to bounded signals, we propose repeating the vector-matrix multiplication after reducing the input strength by a half when a signal saturation is detected. This would guarantee that after a few iterations ($n$) the unbounded signals can be read reliably and then properly rescaled in the digital domain. In this bound management scheme, the effective vector-matrix multiplication on an RPU array can be written as

$$y = \left[W\left[\frac{x}{2^n}\right] + \sigma\right] 2^n \tag{4}$$

with a new effective bound of $2^n|\alpha|$. Note the noise term is also amplified by the same factor; however, the signal to noise ratio remains fixed, only a few percent, for the largest numbers that contribute most in calculation of $softmax$ activations.

In order to test the validity of the proposed noise management (NM) and bound management (BM) techniques, we performed simulations using the RPU-baseline model of Table 1 with and without enabling NM and BM. The summary of these simulations is presented in Figure 3B. When both NM and BM are off, the model using the RPU baseline of Table I, shown as black curve, performs poorly similar to the black curve in Figure 3A. Similarly, turning on either NM or BM alone (as shown by red and blue curves) is not sufficient and the models reach test errors of about 10%. However, when both NM and BM are enabled the model achieves a test error of about 1.7% as shown by the green curve. This is very similar to the model with no analog noise and infinite bounds presented in Figure 3A and shows the success of the noise and bound management techniques. Simply rescaling the numbers in the digital domain these techniques mitigate both the noise and the bound problems inherent to analog computations performed on RPU arrays.

The additional computations introduced in the digital domain due to NM and BM are not significant and can be addressed with a proper digital design. For the NM technique, $\delta_{max}$ needs to be searched in $\boldsymbol{\delta}$ and each element in $\boldsymbol{\delta}$ (and $\boldsymbol{z}$) value needs to be divided and multiplied with this $\delta_{max}$ value. All of these computations require additional $O(M)$ comparison, division and multiplication operations that are performed in the digital domain. However, given that the same circuits need to compute $O(M)$ error signals using the derivative of the activation functions, these additional operations does not change the complexity of the operations that needs to be performed by the digital circuits. Therefore, with proper design these additional operations can be performed with only a slight overhead without causing significant slowdown on the digital circuits. Similarly, BM can be handled in the digital domain by performing $O(N)$ computations only when a signal saturation is detected.

**Sensitivity to Device Variations**

The RPU-baseline model with NM and BM performs reasonable well and achieves a test error of 1.7%, however, this is still above the 0.8% value achieved with a FP-baseline model. In order to identify the dominating factors and layers contributing to this additional classification error, we performed simulations while selectively eliminating various device imperfections from different layers. The summary of these results is shown in Figure 4, where the average test error achieved between 25th and 30th epochs is reported on the y-axis along with an error bar that represents the standard deviation for the same interval. The black data points in Figure 4 corresponds to experiments where device-to-device and cycle-to-cycle variations corresponding to parameters $\Delta w_{min}$, $\Delta w_{min}^+/\Delta w_{min}^-$ and $|w_{ij}|$ are completely eliminated for different layers while the average values are kept unaltered. The model that is free from device variations for all four layers achieves a test error of 1.05%. We note that most of this improvement comes from the convolutional layers as a very similar test error of 1.15% is achieved for the model that does not have device variations for $K_1 \& K_2$, whereas the model without any device variations for fully connected layers $W_3 \& W_4$ remains at 1.3% level. Among the convolutional layers, it is clear that $K_2$ is more influential compared to $K_1$ as test errors of 1.2% or 1.4% are achieved respectively for models with device variations eliminated for $K_2$ or $K_1$. Interestingly, when we repeated similar experiments by eliminating only the device-to-device variation for the imbalance parameter $\Delta w_{min}^+/\Delta w_{min}^-$ from different layers, a very similar trend is observed as shown by the red data points. These results highlights the important of the device imbalance and shows that even a few percent device imbalance can still be harmful while training a network.

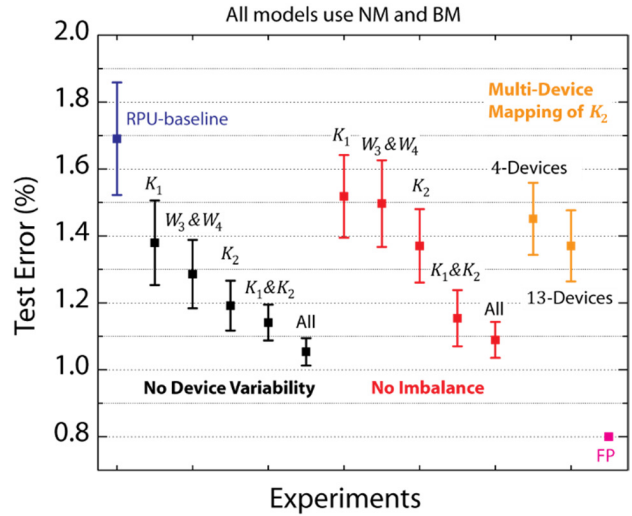

**Figure 4.** Average test error achieved between 25$^{th}$ and 30$^{th}$ epochs for a various RPU models with varying device variations. Black data points corresponds to experiments where device-to-device and cycle-to-cycle variations corresponding to parameters $\Delta w_{min}$, $\Delta w_{min}^+/\Delta w_{min}^-$ and $|w_{ij}|$ are all completely eliminated from different layers. Red data points corresponds to experiments where only the device-to-device variation for the imbalance parameter $\Delta w_{min}^+/\Delta w_{min}^-$ is eliminated from different layers. Green points corresponds to models where multiple RPU devices are mapped for the second convolutional layer $K_2$. RPU-baseline with noise and bound management as well as the FP-baseline models are also included for comparison.

It is clear that the reduction in device variations in some layers can further boost the network performance; however, for realistic technological implementations of the crossbar arrays variations are controlled by fabrication tolerances in a given technology. Therefore, complete or even partial elimination of any device variation is not a realistic option. Instead, in order to get better training results, the effects of the device variations can be mitigated by mapping more than one RPU device per weight, which averages out the device variations and reduces the variability [37]. Here, we propose a flexible multi-device mapping that can be realized in the digital domain by repeating the input signals going to the columns (or rows) of an RPU array, and/or summing (averaging) the results of the output signals generated from the rows (or columns). Since the same signal propagates through many devices and the results are summed on the digital domain, this technique allows to average out device variations in the array without physically hardwiring the lines corresponding to different columns or rows.

To test the validity of this digitally controlled multi-device mapping approach, we performed simulations using models where the mapping of the most influential layer $K_2$ is repeated on 4 or 13 devices. Indeed, this multi-device mapping approach reduces the test error to 1.45% and 1.35% for 4 and 13 device mapping cases, respectively, as shown by the green data points in Figure 4. The number of devices ($\#_d$) used per weight are effectively reduces the device variations by the a factor proportional to $\sqrt{\#_d}$. Note that 13-device mapping of $K_2$ effectively reduces the device variations by a factor of 3.6 at a cost of increase in the array dimensions to $416 \times 401$. However, assuming RPU arrays are fabricated with equal number of columns and rows, multi-device mapping of rectangular matrixes such as $K_2$ does not introduce any operational (or circuit) overhead as long as the mapping fits in the physical dimensions of a square array. In addition, this method allows to flexibly control the number of devices used while mapping different layers and therefore a viable approach to mitigate the effects of device variability for a realistic technological implementations of the crossbar arrays.

**Update Management**

All RPU models presented so far use the stochastic update scheme with a bit length of $BL = 10$ and amplification factors that are equally distributed to the columns and the rows with values $C_x = C_\delta = \sqrt{\eta/(BL\,\Delta w_{min})} = 1.0$. The choice of these values is dictated by the learning rate which is a hyper-parameter of the training algorithm; therefore the hardware should be able to handle any value without imposing any restrictions on it. The learning rate for the stochastic model is the product of four terms; $\Delta w_{min}$, $BL$, $C_x$ and $C_\delta$. Among them $\Delta w_{min}$ corresponds to an incremental conductance change on an RPU device due a single coincidence event; therefore the control of this parameter may be restricted by the hardware. For instance, $\Delta w_{min}$ may be controlled only by shaping the voltage pulses used during the update cycle and hence requires programmable analog circuits. In contrast, the control of $C_x, C_\delta$ and $BL$ is much easier and can be implemented in the digital domain.

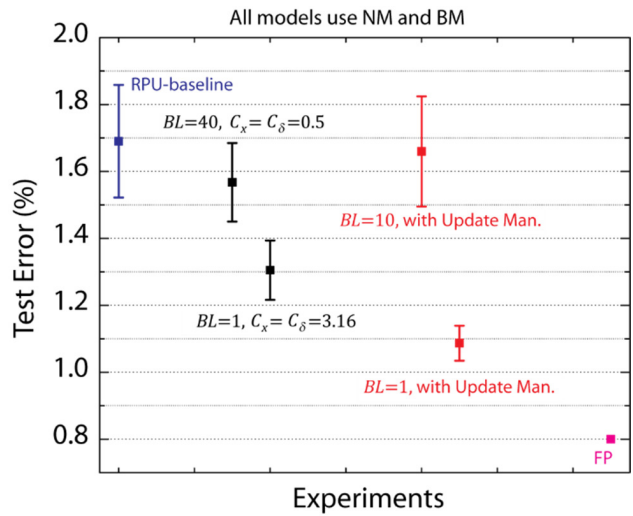

**Figure 5.** Average test error achieved between 25[th] and 30[th] epochs for a various RPU models with varying update schemes. Black data points correspond to updates with amplification factors that are equally distributed to the columns and the rows. Red data points corresponds to models that uses the update management scheme. RPU-baseline with noise and bound management as well as the FP-baseline models are also included for comparison.

To test the effect of $C_x, C_\delta$ and $BL$ on the training accuracy we performed simulations using the RPU-baseline model with the noise and bound management techniques described above while varying those parameters. For all models we used the same fixed learning rate $\eta = 0.01$ and $\Delta w_{min} = 0.001$, and the summary of these results are shown in Figure 5. For the first set of models we varied $BL$, and $\sqrt{\eta/(BL\,\Delta w_{min})}$ value is used equally for the amplification factors $C_x$ and $C_\delta$. Interestingly, increasing $BL$ to 40 did not improve the network performance, whereas reducing it to 1 boosted the performance and a test error of about 1.3% is achieved. These results may be counter intuitive as one may expect the larger $BL$ case to be less noisy and hence would perform better. However, for $BL = 40$ case, the amplification factors are smaller ($C_x = C_\delta = 0.5$) in order to satisfy the same learning rate on average. This reduces the probability of generating a pulse but since there exist longer streams during the update, the average update and the variance do not change. In contrast, for $BL = 1$, the amplifications factors are larger with a value 3.16 and therefore the pulse generation becomes more likely. Indeed for cases when the amplified numbers are larger than unity ($C_x x_i > 1$ or $C_\delta \delta_j > 1$) a single update pulse is generated for sure. This makes the updates more deterministic but with an earlier clipping. Note for $BL = 1$ the weight value can only move by a single $\Delta w_{min}$ per update cycle. However, also note that the convolutional layers $K_1$ and $K_2$ receive 576 and 64 single bit stochastic updates per image due to weight reuse while the fully connected layers $W_3$ and $W_4$ receives only one single bit stochastic update per image. Although the interaction of all of these terms and the tradeoffs are non-trivial, it is clear that the above CNN architecture favors $BL = 1$ whereas the DNN used in Ref [27] favored $BL = 10$. These results emphasize the importance of designing flexible hardware that can control the number of pulses used for the update cycle. We note that this flexibility can be achieved seamlessly for the stochastic update scheme without changing the design considerations for peripheral circuits generating the random pulses.

In addition to $BL$, for the second set of experiments the amplification factors $C_x$ and $C_\delta$ used during the update cycle are also varied to some extend while maintaining the average learning rate fixed. Above models all assume that the same $C_x$ and $C_\delta$ are used during updates; however, it is possible to use arbitrarily larger or smaller $C_x$ and $C_\delta$ as long as the product satisfies $\eta/(BL\ \Delta w_{min})$. In our update management scheme, we use $C_x$ and $C_\delta$ values such that the probability of generating pulses from columns ($\boldsymbol{x}$) and rows ($\boldsymbol{\delta}$) are roughly the same order. This is achieved by rescaling the amplification factors with a ratio $m = \sqrt{\delta_{max}/x_{max}}$, and in this scheme the amplification factors can be written as $C_x = m\sqrt{\eta/(BL\ \Delta w_{min})}$ and $C_\delta = (\frac{1}{m})\sqrt{\eta/(BL\ \Delta w_{min})}$. Although for $BL = 10$ this method did not yield any improvement, for $BL = 1$ the error rate as low as 1.1% is achieved; and hence shows that the proposed update management scheme can yield better training results.

This proposed update scheme does not alter the expected change in the weight value and therefore its benefits may not be obvious. Note that towards the end of training it is very likely the range of numbers in columns ($\boldsymbol{x}$) and rows ($\boldsymbol{\delta}$) are very different; i.e. $\boldsymbol{x}$ have many elements close 1 (or -1) whereas $\boldsymbol{\delta}$ have elements very close to zero ($\boldsymbol{\delta} \ll \boldsymbol{1}$). For this case if the same $C_x$ and $C_\delta$ are used, the updates become row-vise correlated. Although unlikely, generation of a pulse for $\delta_j$ will result in many coincidences along the row $j$, as there are many pulses generated by different columns since many $x_i$ values are close to unity. Our update management scheme eliminates this correlated updates by shifting the probabilities from columns to rows by simple rescaling the numbers used during the update. One may view that this scheme uses rescaled vectors ($m\boldsymbol{x}$ and $\boldsymbol{\delta}/m$) for the updates which are composed of numbers roughly the same order. This update management scheme relies on a simple rescaling of the numbers that are performed in the digital domain; and therefore, does not change the design of the analog circuits needed for the update cycle. The additional computations introduced in the digital domain are not significant either and only require additional $O(M)$ (or $O(N)$) operations similar to the noise management technique.

**Results Summary**

The summary CNN training results of various RPU models that use the above management techniques is shown in Figure 6. When all management techniques are disabled the RPU-baseline model can only achieve test error above 10%. This large error rate is reduced significantly to a value about 1.7% with the noise and bound management techniques, which address the noise issue in the backward cycle and the signal saturation during forward cycle. Additionally when the update management scheme is enabled with a reduced bit length during updates, the model reaches a test error of 1.1%. Finally, the combination of all of those management techniques with the 13-device mapping on the second convolutional brings the model's test error to 0.8% level. The performance of this final RPU model is almost indistinguishable from the FP-baseline model and hence shows the successful application of RPU approach for training CNNs. We note that all those methods can be turned on selectively by simply programing the operations performed on digital circuits; and therefore can be applied to any network architecture beyond CNNs without changing design considerations for realistic technological implementations of the crossbar arrays and analog peripheral circuits.

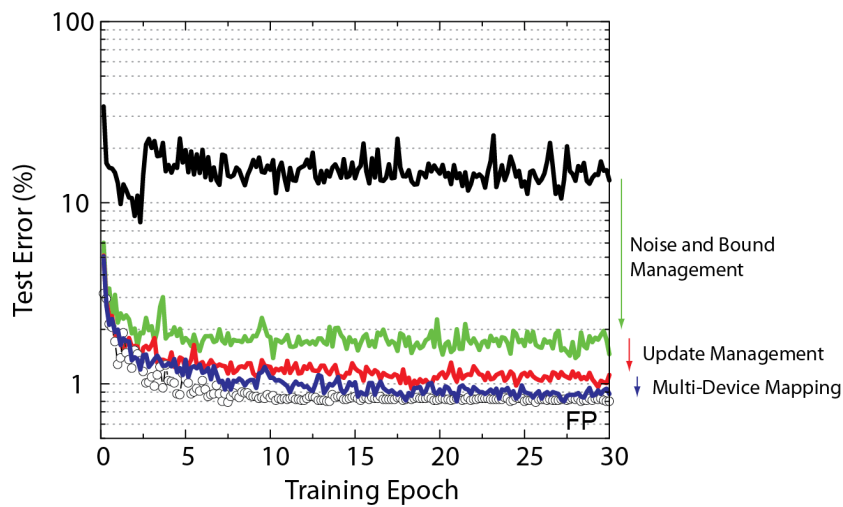

**Figure 6.** Test error of CNN with the MNIST dataset. Open white circles correspond to the model with the training performed using the floating point numbers. Lines with different colors correspond to RPU-baseline model with different management techniques enabled progressively.

## DISCUSSION AND CONCLUSIONS

The application of RPU device concept for training CNNs requires a rearrangement of the kernel parameters and only after this rearrangement the inherent parallelism of the RPU array can be fully utilized for convolutional layers. A single vector operation performed on the RPU array is a constant time $O(1)$ and independent of the array size, however, because of the weight sharing in convolutional layers, the RPU arrays are accessed several times, resulting in a series of vector operations performed on the array for all three cycles. These repeated vector operations introduce interesting challenges and opportunities while training CNNs on a RPU based hardware.

The array sizes, weight sharing factors ($ws$) and the number of multiply and add (MAC) operations performed at different layers for a relative simple but respectable CNN architecture AlexNet [2] are shown in Table 2. This architecture won the large-scale ImageNet competition by a large margin in 2012. We understand that there has been significant progress since 2012 and we only choose AlexNet architecture due to its simplicity and to illustrate interesting possibilities that RPU based hardware enables while designing new network architectures.

**Table 2. Array sizes, weight sharing factors and number of MACs performed for each layer for AlexNet\* [2] architecture**

| Layer | RPU Array Size (Matrix Size) | Weight Sharing Factor ($ws$) | MACs |
|---|---|---|---|
| $K_1$ | $96 \times 363$ | 3025 | $106\,M$ |
| $K_2$ | $256 \times 2400$ | 729 | $448\,M$ |
| $K_3$ | $384 \times 2304$ | 169 | $150\,M$ |
| $K_4$ | $384 \times 3456$ | 169 | $224\,M$ |
| $K_5$ | $256 \times 3456$ | 169 | $150\,M$ |
| $W_6$ | $4096 \times 9216$ | 1 | $38\,M$ |
| $W_7$ | $4096 \times 4096$ | 1 | $17\,M$ |
| $W_8$ | $1000 \times 4096$ | 1 | $4\,M$ |
| | | | $Total\ MACs = 1.14\,G$ |

\*Table assumes the weights that are originally distributed to two GPUs are contained into a single RPU array for each layer

When AlexNet architecture runs on a conventional hardware (such as CPU, GPU or ASIC), the time to process a single image is dictated by the total number of MACs; therefore, the contributions of different layers to the total workload are additive, with $K_2$ consuming about 40% of the workload. The total number of MACs is usually considered as the main metric that scales the training time, and hence, practitioners deliberately construct network architectures while keeping the total number of MACs below a budged to avoid infeasible training times. This constraints the choice of the number of kernels and their dimensions used for each convolutional layer as well as the size of the pooling layers. Assuming a compute bounded system, the time to process a single image on a conventional hardware can be estimated using the ratio of the total number of MACs to the performance metric of the corresponding hardware ($Total\ MACs\ /\ Throughput$).

In contrast to conventional hardware, when the same architecture runs on a RPU based hardware, the time to process a single image is not dictated by the total number of MACs, instead it is dominated by the largest weight reuse factor in the network. For the above example, the operations performed on the first convolutional $K_1$ takes the longest time among all layers because of the large weight reuse factor of $ws = 3025$, although this layer has the smallest array size and only 10% of the total number of MACs. Assuming a RPU based accelerator with many RPU arrays and pipeline stages between them, the average time to process a single image can be estimated as $ws \times t_{meas}$ using values from layer $K_1$, where $t_{meas}$ is the measurement time corresponding to a single vector-matrix multiplication on the RPU array. First of all, this metric emphasizes the constant time operation on RPU arrays as the training time is independent of the array sizes, the number of trainable parameters in the network and the total number of MACs. This would encourage practitioners to use increasing number of kernels with larger dimensions without worrying about training times that is otherwise impossible with conventional hardware. However, the same metric also highlights the importance of $t_{meas}$ and $ws$ for layer $K_1$ that is causing a bottleneck; and therefore, it is desirable to come up with strategies that can reduce both parameters.

In order to reduce $t_{meas}$ we first discuss designing small RPU arrays that can operate faster. It is clear that designing large arrays are beneficial to achieve high degree of parallelism for the vector operations performed on the arrays; however, due to parasitic resistance and capacitance of the transmission lines the largest array size is limited to $4096 \times 4096$ [27]. For an array of this size the measurement time of $t_{meas} = 80ns$ is derived considering the acceptable noise threshold value, which is dominated by the thermal noise on RPU devices. However, for a small array with $512 \times 512$ devices the acceptable noise threshold is not dominated by the thermal noise and hence $t_{meas}$ can be reduced to about $10ns$ for faster computations. It is not desirable to build an accelerator chip all composed of small arrays, as for a small array power and area are dominated by the peripheral circuits (mainly by ADCs); and therefore, a small array has worse power and area efficiency metrics compared to a large array. However, a bimodal design consisting of large and small size arrays achieves better hardware utilization and provides speed advantage while mapping architectures with significantly varying matrix dimensions. While the large arrays are used to map fully connected layers or large convolutional layers, for a convolutional layer such as $K_1$ using the small array would be better a solution that provides a reduction in $t_{meas}$ from $80ns$ to $10ns$.

In order to reduce the weight reuse factor on $K_1$, next we discuss to allocate two (or more) arrays for the first convolutional layer. When more than one array is allocated for the first convolutional layer the network can be forced to learn separate features on different arrays by properly directing the upper (left) and lower (right) portions of the image to separate arrays and by computing the error signals and the updates independently. Not only this allows the network to learn independent features for separate portions of the image, for each array the weight reuse factor is reduced by a factor of 2. This reduces the time to process a single image while making the architecture more expressive. Alternatively, one could also try to synchronize the two arrays by randomly shuffling the portions of the images that are processed by different arrays. This approach would force the network to learn same features on two arrays with same reduction of 2 in the weight reuse factor. These discussed subtle changes in the network architecture do not provide any speed advantage when run on a conventional hardware; and therefore, it highlights the interesting possibilities that a RPU based architecture provides.

In summary, we show that the RPU concept can be applied beyond fully connect networks and the RPU based accelerators are natural fit for training convolutional neural networks as well. These accelerators promise unprecedented speed and power benefits and hardware level parallelism as the number of trainable parameters increase in the network. Because of the constant time operation on RPU arrays, RPU based accelerators provide interesting network architecture choices without increasing the training times. However, all of the benefits of an RPU array are tied to the analog nature of the computations performed on it and therefore introduces some challenges. We show that digitally programmable management techniques are sufficient to eliminate the noise and bound limitations imposed on the array. Furthermore, their combination with the update management and device variability reduction techniques enable a successful application of the RPU concept for training CNNs. All the management techniques discussed in this paper are addressed in the digital domain without changing the design considerations on the array and the supporting analog peripheral circuits. These techniques enable the applicability of RPU approach to wide a variety of networks beyond convolutional or fully connected networks.

# References

[1]  Y. LeCun, Y. Bengio and G. Hinton, "Deep learning," *Nature,* vol. 521, pp. 436-444, 2015.

[2]  A. Krizhevsky, I. Sutskever and G. Hinton, "Imagenet classification with deep convolutional neural networks," *NIPS,* pp. 1097-1105, 2012.

[3]  K. Simonyan and A. Zisserman, "Very Deep Convolutional Networks for Large-Scale Image," *ICLR,* 2015.

[4]  C. Szegedy, W. Liu, Y. Jia, P. Sermanet, S. Reed, D. Anguelov, D. Erhan, V. Vanhoucke and A. Rabinovich, "Going deeper with convolutions," *CVPR,* 2015.

[5]  K. He, X. Zhang, S. Ren and J. Sun, "Delving Deep into Rectifiers: Surpassing Human-Level Performance on ImageNet Classification," in *2015 IEEE International Conference on Computer Vision (ICCV)*, 2015.

[6]  G. Hinton, L. Deng, G. Dahl, A. Mohamed, N. Jaitly, A. Senior, V. Vanhoucke, P. Nguyen, T. Sainath and B. Kingsbury, "Deep neural networks for acoustic modeling in speech recognition: The shared views of four research groups," *IEEE Signal Processing Magazine,* pp. 82-97, 2012.

[7]  R. Collobert, J. Weston, L. Bottou, M. Karlen, K. Kavukcuoglu and P. Kuksa, "Natural Language Processing (Almost) from Scratch," *Journal of Machine Learning Research,* vol. 12, pp. 2493-2537, 2012.

[8]  Q. Le, M. Ranzato, R. Monga, M. Devin, K. Chen, G. Corrado, J. Dean and A. Ng, "Building high-level features using large scale unsupervised learning," *International Conference on Machine Learning,* 2012.

[9] S. Gupta, W. Zhang and F. Wang, "Model Accuracy and Runtime Tradeoff in Distributed Deep Learning: A Systematic Study," *IEDM,* 2016.

[10] J. Dean, G. Corrado, R. Monga, K. Chen, M. Devin, Q. Le, M. Mao, M. Ranzato, A. Senior, P. Tucker, K. Yang and A. Ng, "Large scale distributed deep networks," in *NIPS'12*, 2012.

[11] A. Coates, B. Huval, T. Wang, D. Wu and A. Ng, "Deep learning with COTS HPC systems," *ICML,* 2013.

[12] R. Wu, S. Yan, Y. Shan, Q. Dang and G. Sun, "Deep Image: Scaling up Image Recognition," 2015.

[13] S. Gupta, A. Agrawal, K. Gopalakrishnan and P. Narayanan, "Deep Learning with Limited Numerical Precision," 2015.

[14] Y. Chen, T. Luo, S. Liu, S. Zhang, L. He, J. Wang, L. Li, T. Chen, Z. Xu, N. Sun and O. Temam, "DaDianNao : A Machine-Learning Supercomputer," *47th Annual IEEE/ACM International Symposium on Microarchitecture,* pp. 609-622, 2014.

[15] J. Emer, V. Sze and Y. Che, "Tutorial on Hardware Architectures for Deep Neural Networks," in *IEEE/ACM International Symposium on Microarchitecture (MICRO-49)*, 2016.

[16] Y. Arima, K. Mashiko, K. Okada, T. Yamada, A. Maeda, H. Notani, H. Kondoh and S. Kayano, "A 336-neuron, 28 K-synapse, self-learning neural network chip with branch-neuron-unit architecture," *IEEE Journal of Solid-State Circuits,* vol. 26, p. 1991, 1637-1644.

[17] C. Lehmann, M. Viredaz and F. Blayo, "A generic systolic array building block for neural networks with on-chip learning," *IEEE Transactions on Neural Networks,* vol. 4, pp. 400-407, 1993.

[18] G. Burr, R. Shelby, A. Sebastian, S. Kim, S. Kim, S. Sidler, K. Virwani, M. Ishii, P. Narayanan, A. Fumarola, L. Sanches, I. Boybat, M. Le Gallo, K. Moon, J. Woo, H. Hwang and Y. Leblebici, "Neuromorphic computing using non-volatile memory," *Advances in Physics : X,* pp. 89-124, 2017.

[19] D. Kuzum, S. Yu and H. Wong, "Synaptic electronics: materials, devices and applications," *Nanotechnology,* vol. 24, 2013.

[20] P. Chi, S. Li, C. Xu, T. Zhang, J. Zhao, Y. Liu, Y. Wang and Y. Xie, "PRIME: A novel processing-in-memory architecture for neural network computation in ReRAM based main memory," in *ISCA*, 2016.

[21] M. Prezioso, F. Merrikh-Bayat, B. Hoskins, G. Adam, A. Likharev and D. Strukov, "Training and operation of an integrated neuromorphic network based on metal-oxide memristors," *Nature,* pp. 61-64, 2015.

[22] E. Merced-Grafals, N. Davila, N. Ge, S. Williams and J. Strachan, "Repeatable, accurate, and high speed multi-level programming of memristor 1T1R arrays for power efficient analog computing applications," *Nanotechnology,* 2016.

[23] D. Soudry, D. Di Castro, A. Gal, A. Kolodny and S. Kvatinsky, "Memristor-Based Multilayer Neural Networks With Online Gradient Descent Training," *IEEE Transactions on Neural Networks and Learning Systems ,* 2015.

[24] Z. Xu, A. Mohanty, P. Chen, D. Kadetotad, B. Lin, J. Ye, S. Vrudhula, S. Yu, J. Seo and Y. Cao, "Parallel Programming of Resistive Cross-point Array for Synaptic Plasticity," *Procedia Computer Science,* vol. 41, pp. 126-133, 2014.

[25] G. Burr, P. Narayanan, R. Shelby, S. Sidler, I. Boybat, C. di Nolfo and Y. Leblebici, "Large-scale neural networks implemented with non-volatile memory as the synaptic weight element: Comparative performance analysis (accuracy, speed, and power)," *IEDM (International Electron Devices Meeting),* 2015.

[26] J. Seo, B. Lin, M. Kim, P. Chen, D. Kadetotad, Z. Xu, A. Mohanty, S. Vrudhula, S. Yu, J. Ye and Y. Cao, "On-Chip Sparse Learning Acceleration With CMOS and Resistive Synaptic Devices," *IEEE Transactions on Nanotechnology,* 2015.

[27] T. Gokmen and Y. Vlasov, "Acceleration of Deep Neural Network Training with Resistive Cross-Point Devices," *Frontiers in Neuroscience,* 2016.

[28] S. Agrawal, T. Quach, O. Parekh, A. Hsia, E. DeBenedictis, C. James, M. Marinella and J. Aimone, "Energy Scaling Advantages of Resistive Memory Crossbar Computation and Its Application to Sparse Coding," *Frontiers in Neuroscience,* 2016.

[29] E. Fuller, F. El Gabaly, F. Leonard, S. Agarwal, S. Plimpton, R. Jacobs-Gedrim, C. James, M. Marinella and A. Talin, "Li-Ion Synaptic Transistor for Low Power Analog Computing," *Advanced Science News,* vol. 29, 2017.

[30] S. Agrawal, S. Plimpton, D. Hughart, A. Hsia, I. Richter, J. Cox, C. James and M. Marinella, "Resistive Memory Device Requirements for a Neural Network Accelerator," *IJCNN,* 2016.

[31] D. Rumelhart, G. Hinton and R. Williams, "Learning representations by back-propagating errors," *Nature,* vol. 323, pp. 533-536, 1986.

[32] K. Steinbuch, "Die Lernmatrix," *Kybernetik,* 1961.

[33] Y. LeCun, L. Bottou, Y. Bengio and P. Haffner, "Gradient-Based Learning Applied to Document Recognition," *Proceedings of the IEEE,* vol. 86, no. 11, pp. 2278-2324, 1988.

[34] B. Gaines, "Stochastic computing," in *Proceedings of the AFIPS Spring Joint Computer Conference*, 1967.

[35] W. Poppelbaum, C. Afuso and J. Esch, "Stochastic computing elements and systems," in *Proceedings of the AFIPS Fall Joint Computer Conference*, 1967.

[36] C. Merkel and D. Kudithipudi, "A stochastic learning algorithm for neuromemristive systems," in *27th IEEE International System-on-Chip Conference (SOCC)*, 2014.

[37] P. Chen, D. Kadetotad, Z. Xu, A. Mohanty, B. Lin, J. Ye, S. Vrudhula, J. Seo, Y. Cao and S. Yu, "Technology-design co-optimization of resistive cross-point array for accelerating learning algorithms on chip," in *DATE*, 2015.